\documentclass[10pt,twocolumn,letterpaper]{article}
\usepackage[accsupp]{axessibility}  
\usepackage{iccv}
\usepackage{times}
\usepackage{epsfig}
\usepackage{eso-pic}
\usepackage{graphicx}
\usepackage{amsmath}
\usepackage{amssymb}
\usepackage{float}
 \usepackage{pifont}
\usepackage{xcolor}
\usepackage{soul}
\usepackage{bbding}
\usepackage{booktabs}
\usepackage[normalem]{ulem}
\usepackage{multirow}
\newcommand{\cmark}{\ding{51}}%
\newcommand{\xmark}{\ding{55}}%
\usepackage{enumitem}
\usepackage[breaklinks=true,bookmarks=false]{hyperref}

\usepackage[capitalize]{cleveref}
\crefname{section}{Sec.}{Secs.}
\Crefname{section}{Section}{Sections}
\Crefname{table}{Table}{Tables}
\crefname{table}{Tab.}{Tabs.}
\usepackage[capitalize]{cleveref}
\crefname{section}{Sec.}{Secs.}
\Crefname{section}{Section}{Sections}
\Crefname{table}{Table}{Tables}
\crefname{table}{Tab.}{Tabs.}

\usepackage[breaklinks=true,bookmarks=false]{hyperref}

\iccvfinalcopy 

\def\iccvPaperID{6735} 

\ificcvfinal\fi

\begin{document}

\title{DiffCloth: Diffusion Based Garment Synthesis and Manipulation via\\ Structural Cross-modal Semantic Alignment}

\author{
Xujie Zhang{$^{1}$}\thanks{}
\quad Binbin Yang{$^{2*}$}
\quad Michael C. Kampffmeyer{$^{4}$}
\quad Wenqing Zhang{$^{1}$} \\ 
\quad Shiyue Zhang{$^{1}$}
\quad Guansong Lu{$^{3}$}
\quad Liang Lin{$^{2}$}
\quad Hang Xu{$^{3}$}
\quad Xiaodan Liang{$^{1,5}$$^\dag$} \\
{\normalsize{$^{1}$}Shenzhen Campus of Sun Yat-Sen University \quad {$^{2}$}Sun Yat-Sen University \quad} \\{\normalsize{$^{3}$}Huawei Noah's Ark Lab\quad
{$^{4}$}UiT The Arctic University of Norway\quad
{$^{5}$}MBZUAI}\\
\small{\tt{\{zhangxj59,yangbb3,zhangwq76,zhangshy223\}@mail2.sysu.edu.cn,michael.c.kampffmeyer@uit.no}}\\ 
\small{\tt{luguansong@huawei.com,linliang@ieee.org,chromexbjxh@gmail.com,xdliang328@gmail.com}}
\vspace{-2mm}
}


\twocolumn[{%
\maketitle
\ificcvfinal\thispagestyle{empty}\fi
\vspace{-10mm}
\begin{minipage}{1\textwidth}
    \begin{figure}[H]
        \centering
        \includegraphics[width=1\textwidth]{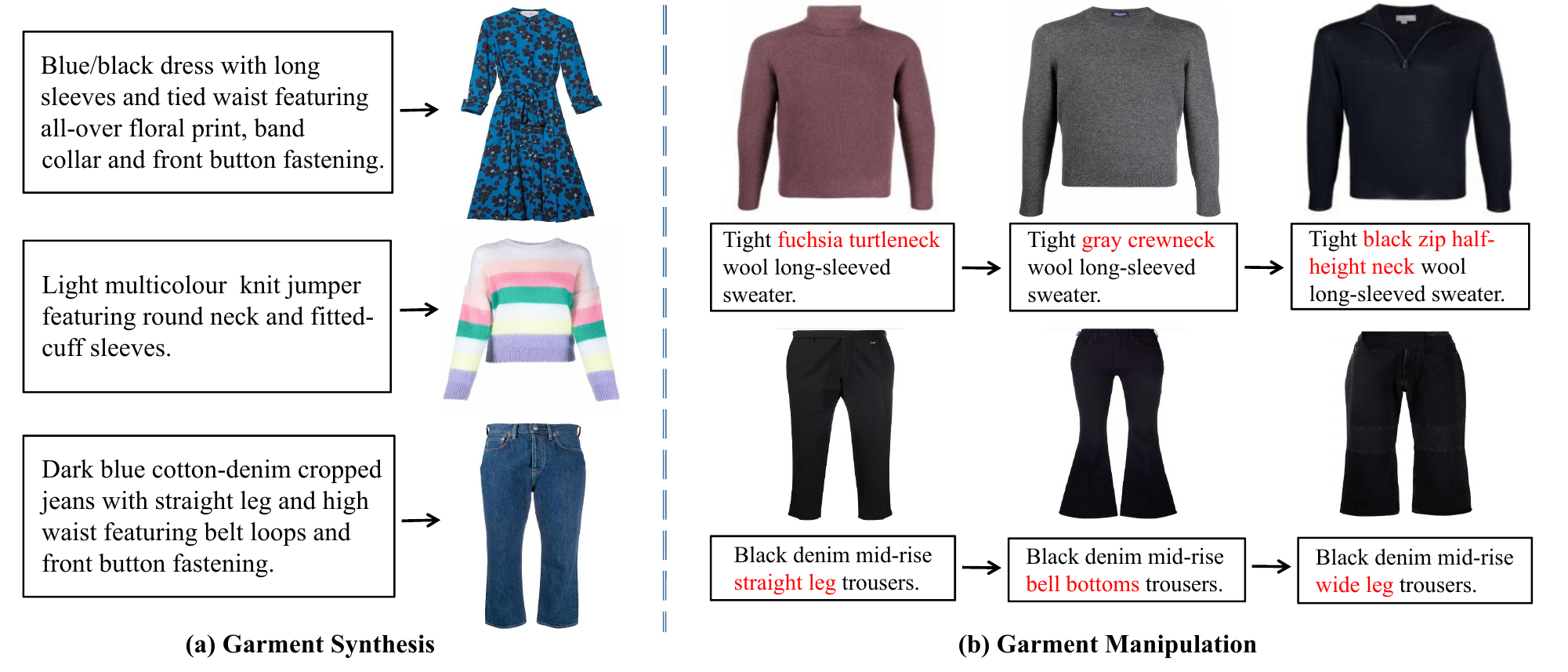}
        \caption{
        Results of our proposed DiffCloth. DiffCloth is able to produce garments with part-level semantics well-aligned to the prompt and allows for precise manipulation of the generated results by simply modifying the text description.
        }
        \vspace{5mm}
        \label{fig:teaser}
    \end{figure}
\end{minipage}
}]

{
  \renewcommand{\thefootnote}%
    {\fnsymbol{footnote}}
  \footnotetext[1]{Equal contribution. $^\dag$Corresponding author.}
}

\begin{abstract}
Cross-modal garment synthesis and manipulation will significantly benefit the way fashion designers generate garments and modify their designs via flexible linguistic interfaces. However, despite the significant progress that has been made in generic image synthesis using diffusion models, producing garment images with garment part level semantics that are well aligned with input text prompts and then flexibly manipulating the generated results still remains a problem.
Current approaches follow the general text-to-image paradigm and mine cross-modal relations via simple cross-attention modules, neglecting the structural correspondence between visual and textual representations in the fashion design domain. In this work, we instead introduce DiffCloth, a diffusion-based pipeline for cross-modal garment synthesis and manipulation, which empowers diffusion models with flexible compositionality in the fashion domain by structurally aligning the cross-modal semantics. Specifically, we formulate the part-level cross-modal alignment as a bipartite matching problem between the linguistic Attribute-Phrases (AP) and the visual garment parts which are obtained via constituency parsing and semantic segmentation, respectively.
To mitigate the issue of attribute confusion, we further propose a semantic-bundled cross-attention to preserve the spatial structure similarities between the attention maps of attribute adjectives and part nouns in each AP. Moreover, DiffCloth allows for manipulation of the generated results by simply replacing APs in the text prompts. The manipulation-irrelevant regions are recognized by blended masks obtained from the bundled attention maps of the APs and kept unchanged. Extensive experiments on the CM-Fashion benchmark demonstrate that DiffCloth both yields state-of-the-art garment synthesis results by leveraging the inherent structural information and supports flexible manipulation with region consistency.
\end{abstract}

\section{Introduction}

Leveraging artificial intelligence to generate and alter garment images based on control signals from a variety of modalities has the potential to revolutionize the fashion design process.
Particularly, cross-modal garment synthesis~\cite{dong2017i2t2i,hong2018inferring,huang2018introduction,lao2019dual,li2019object,reed2016generative,yuan2018text} and manipulation by linguistic interfaces have gradually attracted increasing attention from the academic community.
Unfortunately, the visual semantics in the fashion domain are different from those in generic image generation tasks due to its inherent structural property, \emph{e.g.} each type of garment has a distinct shape and can be partitioned into several garment parts.
However, existing work~\cite{dong2017i2t2i,hong2018inferring,huang2018introduction,lao2019dual,li2019object,reed2016generative,yuan2018text} on cross-modal garment synthesis are primarily built on two-stage pipelines of generic generative transformers and ignore the structural correspondences between the garment images and the input text prompts. This leads to imprecise cross-modal semantic alignment and poor semantic compositionality.

\begin{figure}[tbp]
\centering 
\includegraphics[width=\linewidth]{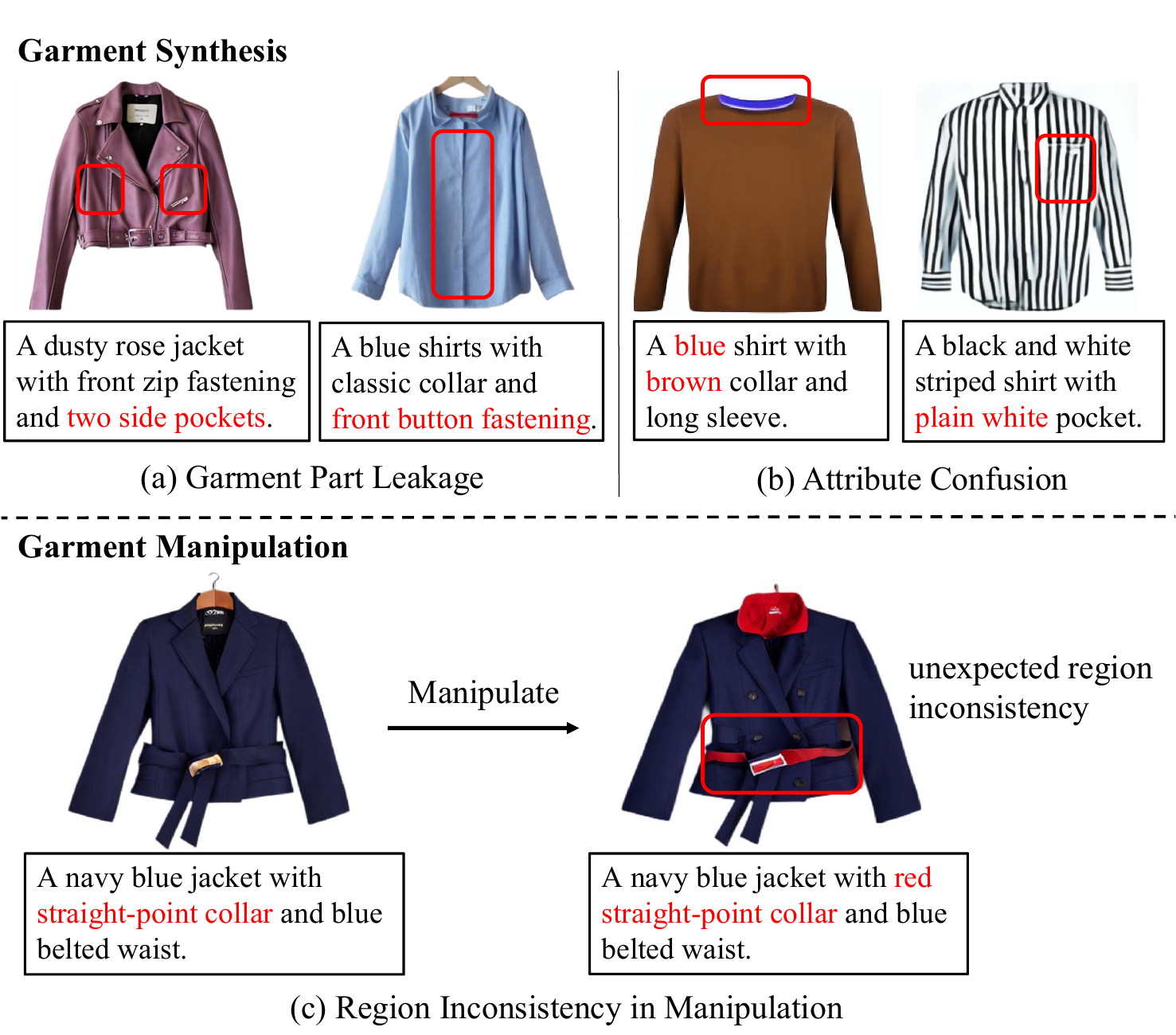} 
\vspace{-6mm}
\caption{Three typical issues of garment synthesis and manipulation. (a) Garment Part Leakage: one or more of the garment parts described in the prompt are not accurately generated; (b) Attribute Confusion: the attributes and garment parts are wrongly paired or some attributes are ignored; (c) Region Inconsistency in Manipulation: the manipulation-irrelevant regions are carelessly modified.}
\vspace{-6mm}
\label{fig:motivation} 
\end{figure}

Given the recent success of diffusion models~\cite{DBLP:conf/icml/NicholDRSMMSC22,DBLP:journals/corr/abs-2204-06125,DBLP:conf/cvpr/RombachBLEO22,DBLP:journals/corr/abs-2205-11487}, which provide flexible control of the generative process through guidance mechanisms, departing from prior approaches and leveraging diffusion models appears a natural approach. However, we observe the following two semantic issues when applying state-of-the-art text-based image generation models to the fashion domain: 1) Garment Part Leakage, where one or more of the garment parts described in the prompt are not actually generated in the image; and 2) Attribute Confusion, where the attributes and the garment parts are wrongly paired or some attributes are ignored in the generated image. Examples of the aforementioned issues are provided in \cref{fig:motivation}. In Fig. \ref{fig:motivation}(a), examples of garment part leakage are provided, where the model fails to generate the pockets in the dusty rose jacket and the button fastening in the blue shirt. In Fig. \ref{fig:motivation}(b), examples of attribute confusion are provided, where the color attributes `blue' and `brown' bind to the incorrect garment parts and the `plain white' attribute is missing in the striped shirt.

To solve the above issues, we propose \textbf{DiffCloth}, a diffusion model with structural semantic consensus guidance to achieve accurate fine-grained part-level semantic alignment. To be specific, a semantic segmentor is trained to explore the visual structure and divide the visual garment into part-level images, \emph{e.g.}, sleeves, body piece, hood, etc. Additionally, a constituency parsing tree is leveraged as a linguistic structural parser to extract the collection of Attribute-Phrases. By formulating the cross-modal semantic alignment as a bipartite matching problem between these two sets of semantic components, we introduce a Hungarian matching loss as the summation of CLIP-similarities~\cite{DBLP:conf/icml/RadfordKHRGASAM21} between the part-level images and the Attribute-Phrases. This Hungarian matching loss can be used to guide the diffusion model to achieve structural consensus across images and text. Furthermore, we propose a semantic-bundled cross-attention module to avoid the aforementioned attribute confusion issue. Specifically, we observe that the attention maps of the attribute adjective and the garment part nouns are different when attribute confusion occurs while they share similar spatial structures when attributes are matched to the correct garment parts. Hence, we propose to preserve the spatial structure similarity between the attribute adjective and the garment part subject in the cross-attention module by a semantic-bundled loss, which aims to minimize the Jensen-Shannon divergence~\cite{DBLP:conf/isit/FugledeT04} between these two maps. This semantic-bundle loss is also utilized to guide the sampling process of DiffCloth.

In order to further allow easy manipulation of the generated images, DiffCloth introduces a mechanism to manipulate input images based solely on changes in the input text prompt and, unlike prior approaches ~\cite{DBLP:conf/cvpr/AvrahamiLF22,ChenlinMeng2021SDEditIS}, does not require explicit masking of the areas that should be changed. By injecting the cross-attention maps during the diffusion steps, DiffCloth can automatically find which pixels should be attended to and should be modified. When for instance changing the attribute, \emph{e.g.} ``long sleeve'' $\rightarrow$ ``short sleeve'', only the cross-attention maps of the bundled Attribute-Phrase need to be changed and the attention maps of other textual tokens can be frozen. Moreover, we propose a consistency loss to prevent irrelevant content from being carelessly edited. An example of unexpected region inconsistency is given in Fig.\ref{fig:motivation} (c), where the blue belt is wrongly modified to a red one. The consistency loss is further designed to preserve the pixel-level consistency of the exclusive area indicated by the attention map of the changed tokens. Comprehensive experiments on the CM-Fashion benchmark demonstrate that DiffCloth yields state-of-the-art generation results in garment synthesis and further supports flexible manipulation by editing the text prompt in a user-friendly manner.

Our main contributions are summarized as follows:\vspace{-2mm}
\begin{itemize}
    \item We propose a structural semantic consensus guidance to address the structural semantic alignment across visual garments and linguistic attribute-phrases as a bipartite matching problem via the Hungarian algorithm.
    \vspace{-6mm}
    \item We propose a new semantic-bundled cross-attention, which encourages spatial structure similarity between the cross-attention maps of attributes and part subjects, to alleviate attribute confusion issues.
    \vspace{-2mm}
    \item We introduce a region consistency mechanism to prevent irrelevant content from being modified during garment manipulation.
    \vspace{-2mm}
    \item Extensive experiments on the CM-Fashion benchmark verify the superiority of DiffCloth, particularly in terms of accurate text-image alignment for both garment synthesis and manipulation.
\end{itemize}

 \begin{figure*}
  \centering
  \includegraphics[width=0.98\hsize]{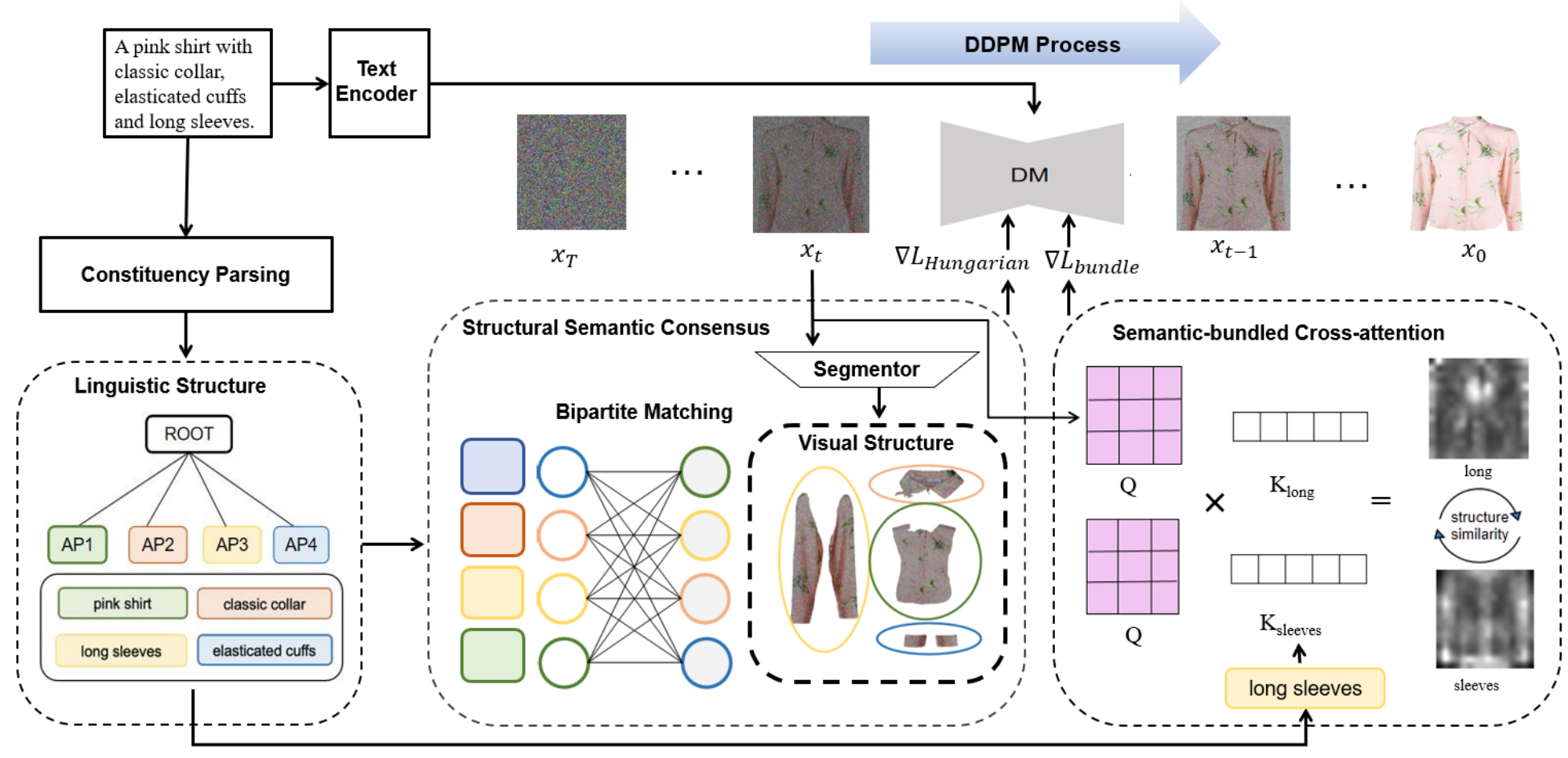}

  \caption{Overview of  DiffCloth. During the diffusion step, we leverage constituency parsing to extract the text structure and obtain a tree of all attribute- phrases (APs). Given this structure information, the structural semantic consensus partitions the garment images using a segmentor into multiple visual parts, which are then matched with the APs using a bipartite matching to get structural semantic alignment. This generates the  $L_{Hungarian}$ loss. Similarly, to preserve structure similarity between the attention maps of the attribute adjectives and the corresponding garment part subjects we introduce semantic-bundled cross-attention, which addresses the attribute confusion issue via the $L_{bundle}$ loss. More specifically, query $Q$ is obtained from the visual representation $X_t$, while keys $K$ are computed for each word. $L_{bundle}$ then aims to encouraging similar attention maps for each AP. Finally, the losses are used to refine the feature representation of the diffusion model at each step.}
  \vspace{-4.5mm}
\label{fig:overview}
\end{figure*}
\section{Related work}
\noindent\textbf{Text-guided image synthesis.} Early works explored text-guided image synthesis in the context of GANs~\cite{DBLP:conf/cvpr/Tao00JBX22,DBLP:conf/cvpr/XuZHZGH018,DBLP:conf/bmvc/YeYTSJ21,DBLP:conf/cvpr/0010KBLY21,DBLP:conf/cvpr/ZhuP0019} or VQVAE~\cite{oord2017neural}.
However, more recent research has demonstrated impressive results using large-scale auto-regressive models~\cite{DBLP:conf/icml/RameshPGGVRCS21,DBLP:journals/corr/abs-2206-10789} and diffusion models~\cite{DBLP:conf/icml/NicholDRSMMSC22,DBLP:journals/corr/abs-2204-06125,DBLP:conf/cvpr/RombachBLEO22,DBLP:journals/corr/abs-2205-11487}. Stable Diffusion~\cite{DBLP:conf/cvpr/RombachBLEO22} proposes to encode an image with an autoencoder and then leverage a diffusion model to generate continuous feature maps in the latent space. Imagen~\cite{DBLP:journals/corr/abs-2205-11487} addresses the importance of language understanding by using a frozen T5~\cite{ColinRaffel2019ExploringTL} encoder, a dedicated large language model. However, generating images that faithfully align with the input prompt remains challenging. To enforce heavier reliance on the text, classifier-free guidance~\cite{DBLP:journals/corr/abs-2207-12598,DBLP:conf/icml/NicholDRSMMSC22,DBLP:journals/corr/abs-2205-11487} allows extrapolating text-driven gradients to better guide the generation by strengthening the reliance on the text. Despite this, the semantic flaws of text-to-image models still exist. Recent work has begun to address this issue, such as Composable Diffusion models~\cite{DBLP:conf/eccv/LiuLDTT22}, which compose multiple outputs of a pre-trained diffusion model. Each output is tasked with capturing different image components which are then joined using compositional operators to attain a unified image. StructureDiffusion~\cite{DBLP:journals/corr/abs-2212-05032} and Attend-and-Excite~\cite{DBLP:journals/corr/abs-2301-13826} optimize the attention map calculation for better image generation.
However, these attempts still fall short of generating garment images with fine-grained compliance with the input prompts as the structural correspondences between garment representations of the two modalities are often ignored. In this work, we strive to achieve part-level cross-modal semantic alignment by aligning those visual and linguistic structured representations in a fine-grained manner.

\noindent\textbf{Image Manipulation with Generative Models.}
A number of techniques~\cite{DBLP:conf/cvpr/AvrahamiLF22,DBLP:journals/corr/abs-2210-09276,DBLP:journals/corr/abs-2208-12242,DBLP:journals/corr/abs-2212-06909} have been developed based on diffusion models to enable editing, personalization and inversion to token space. Dreambooth~\cite{DBLP:journals/corr/abs-2208-12242} and Imagic~\cite{DBLP:journals/corr/abs-2210-09276} involve fine-tuning of the generative models. ImagenEditor~\cite{DBLP:journals/corr/abs-2212-06909} frames editing as text-guided image inpainting, and involves user specified masks. Blended diffusion~\cite{DBLP:conf/cvpr/AvrahamiLF22} provides a clip-guided mask-based editing method. However, the mask provided by the user is often not accurate enough, and there will be disharmony in the editing boundary. 
More recently, Prompt-to-Prompt~\cite{DBLP:journals/corr/abs-2208-01626} explored mask-free image editing through the interaction of attention maps. However, the manipulation results often affect content that is irrelevant to the modification, leading to unsatisfactory results.
In this paper, we explore an attention-based garment manipulation method by injecting the attention maps of the target Attribute-Phrase (AP) while keeping other regions unchanged using a mask that blends the attention maps. 


\section{Methodology}
Our proposed DiffCloth is built on Stable Diffusion~\cite{DBLP:conf/cvpr/RombachBLEO22}, which we briefly review in \cref{sec:sdm}. We then introduce our structural semantic consensus guidance in \cref{sec:structural}, which addresses the problem of garment part leakage. Our semantic-bundled cross-attention mechanism is then presented in \cref{sec:semantic bundle} in order to avoid the confusion between attributes before we present our garment manipulation in \cref{sec:manipulation}. An overview of DiffCloth is provided in Fig.~\ref{fig:overview}.

\subsection{Preparatory}
\label{sec:sdm}
\textbf{Stable Diffusion.} Our proposed DiffCloth is built on Stable Diffusion~\cite{DBLP:conf/cvpr/RombachBLEO22}, which consists of an autoencoder model and a diffusion model. The autoencoder is trained to encode an image $x_0$ as lower-resolution latent maps $z_0$ for efficient diffusion training:
\begin{equation}
    L_{AE} = \Vert x_0 - \text{Dec}(\text{Enc}(x_0))\Vert^2,
\end{equation}
where $L_{AE}$ is the reconstruction loss for training the encoder $\text{Enc}$ and decoder $\text{Dec}$. $z_0 = \text{Enc}(x_0)$ and $x_0$ can be approximately reconstructed by $\text{Dec}(z_0)$. The diffusion model contains two stages: a diffusion and a denoising stage. In the diffusion stage, $z_0$ is gradually transformed into a normal distribution by gradually adding noise for $T$ steps following the Gaussian transition $q(z_t | z_{t-1})$:
\begin{equation}
    q(z_t | z_{t-1}) = \mathcal{N}(z_t; \sqrt{1-\beta_t}z_{t-1}, \beta_t \mathbf{I}),
\end{equation}
where $\beta$ denotes the noise scale, $\mathbf{I}$ is the identity matrix, and $z_t$ is the latent of the timestep $t$ .

By optimizing a noise estimator $\epsilon_\theta$, the model is trained to reverse the diffusion process and generate images from random noise by optimizing the loss $L_{\text{DM}}$:

\begin{equation}
\label{eq:DM}
    L_{\text{DM}} = \mathbb{E}_{t, z_0, \epsilon}\left[\Vert\epsilon_{\theta}(z_t) - \epsilon\Vert^2\right].
\end{equation} A synthesized image $x^*_0$ is obtained by denosing noise $x_T$ for $T$ steps and decoding it using the decoder $x^*_0 = \text{Dec}(z^*_0)$.

DiffCloth is trained on the garment images by optimizing \cref{eq:DM} and sampled using the guidance from our proposed structure semantic consensus and semantic-bundled losses, which will be detailed in the following sections.

\subsection{Structural Semantic Consensus Guidance}
\label{sec:structural}

Our structural semantic consensus guidance is based on the intuition that there are structural similarities between visual and textual representations in cross-modal garment synthesis. 
As shown in Fig.~\ref{fig:overview}, a segmentor trained on noisy inputs can be used to partition garment images into multiple visual parts that adhere to the standard structural patterns used by humans in garment design.\footnote{{More details are provided in \cref{sec:EXP}.}}
The visual structured components can be denoted as:
\begin{equation}
\label{eq:visual_structure}
\mathbf{V} = [V_{full}, V_1, V_2, ..., V_m],
\end{equation}
where $V_{full}$ denotes the full garment image and $V_i$ is the $i^{th}$ part image of $V_{full}$ indicated by the mask $M_i$, \emph{e.g.}, sleeves, body piece, hood.

Similarly, we can obtain the text structure by leveraging constituency parsing to extract a tree of all Attribute-Phrases (APs), which are crucial for depicting the semantic components of a garment image:
\begin{equation}
\mathbf{W} = [W_{full}, W_1, W_2, ..., W_m],
\end{equation}
where $W_{full}$ denotes the full prompt and $W_i$ is the $i^{th}$ linguistic AP in the tree structure \emph{e.g.}, `blue sweater', `classic hood', `long sleeves', where meaningless conjunctions, \emph{e.g.}, `and', `with' are omitted.

\textbf{Bipartite Matching.} In order to generate garment images with part-level consensus between these two collections of visual and linguistic components, we formulate the cross-modal semantic alignment
as a set-to-set bipartite matching problem. 
Our objective is to find a permutation, $\sigma$, of the set of $m$ semantic components, which minimizes the pair-wise semantic matching loss $L_{match}(V_i, W_{\hat{\sigma}(i)})$:
\begin{equation}
    \hat{\sigma} = \underset{\sigma}{\arg\min}\sum_{i}^m L_{match}(V_i, W_{\hat{\sigma}(i)}),
\end{equation}
where $L_{match}(V_i, W_{\hat{\sigma}(i)}) = \text{CLIP}(V_i, W_{\hat{\sigma}(i)})$ and $\text{CLIP}(\cdot, \cdot)$ denotes the CLIP similarity~\cite{DBLP:conf/icml/RadfordKHRGASAM21}. 
The optimal matching is obtained by using the Hungarian algorithm~\cite{HaroldWKuhn1955TheHM}. Further, we define our Hungarian matching loss to compute the hierarchical structure alignment score on $\mathbf{V}$ and $\mathbf{W}$ by calculating the part-level and image-level alignment scores:
\begin{equation}
\small
    L_\text{Hungarian}(\mathbf{V}, \mathbf{W}) = \sum_{i}^m \text{CLIP}(V_i, W_{\hat{\sigma}(i)}) + \text{CLIP}(V_{full}, W_{full}).
\end{equation}

The latent code $z_t$ in the $t^{th}$ denoising step of the diffusion model is then refined via the structural semantic consensus guidance:

\begin{equation}
\label{eq:semantic guidance}
    \hat{z_t} \leftarrow z_t + \alpha\cdot\nabla_{z_t}L_{Hungarian}(\phi(z_t), \mathbf{W}),
\end{equation}
where $\phi(z_t)$ denotes the collection of visual structured components for the decoded image corresponding to $z_t$.

\begin{figure}[tbp]
\centering 
\includegraphics[width=\linewidth]{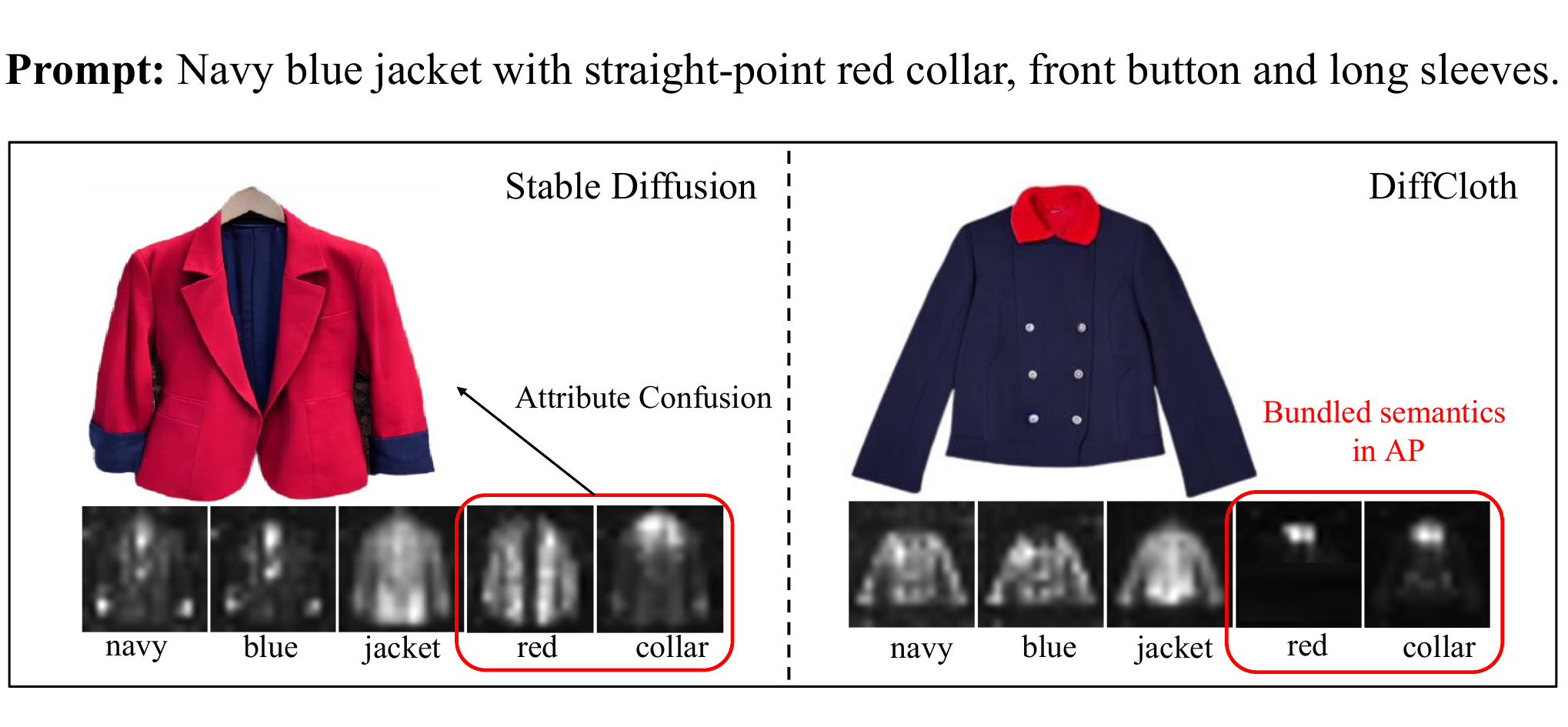} 
\caption{Visualization of the cross-attention of Stable Diffusion and our DiffCloth.}

\label{fig:motiv_bundle} 
\end{figure}

\subsection{Semantic-bundled Cross-attention}
\label{sec:semantic bundle}
Current text-to-image diffusion models, such as Stable Diffusion, have demonstrated that cross-attention between prompt tokens and visual feature maps results in coarse semantic alignment. However, for complex garment descriptions, a phenomenon of `attribute confusion' arises, which can severely impact the reliability of fashion generators. Specifically, attributes and garment parts may be wrongly paired and some attributes may be ignored in the generated image, resulting in imprecise and unsatisfactory generated garments. An example of this is provided in \cref{fig:motiv_bundle}, where the output image is a `Red jacket with blue cuffs.' while the input prompt is `Navy blue jacket with red collar.' To reveal the underlying reason for the incorrect attribution of `red', we visualize the cross-attention maps between the visual tokens and the linguistic tokens in \cref{fig:motiv_bundle}. It can be observed from the results given by Stable Diffusion that the attention map of `red' is spatially similar to that of `jacket' rather than `collar', which leads to the unexpected mismatched attention regions for the Attribute-Phrase pair `red collar'.
\begin{figure}[tbp]
\centering 
\includegraphics[width=\linewidth]{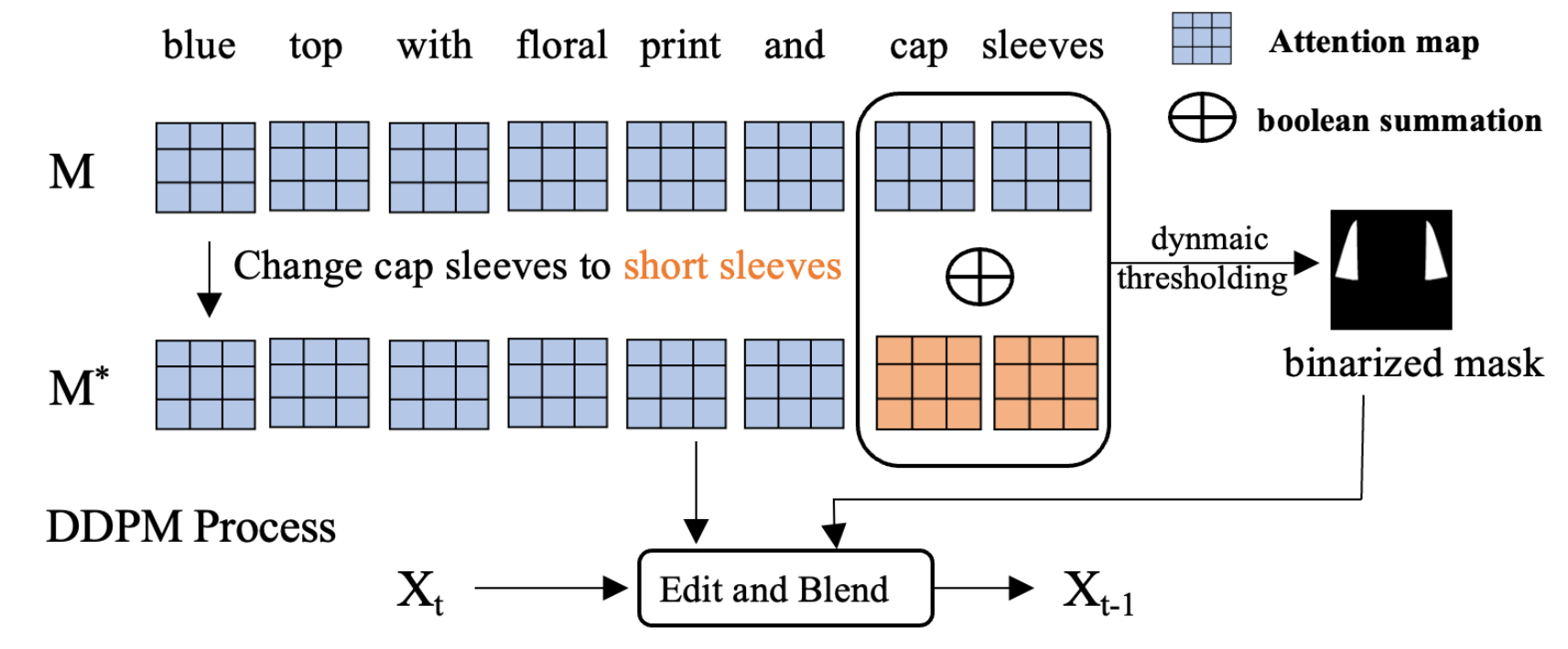} 
\caption{Overview of the garment manipulation pipeline.}
\label{fig:edit_process} 
\end{figure}
To address this issue, we propose a semantic-bundled cross-attention mechanism that leverages a semantic-bundled loss $L_{bundle}$ to preserve the spatial structure similarity between the attention maps of the attribute adjectives and the garment part subject. Formally, given an input prompt $W_{full}$, we first obtain the collection of attribute-phrases $\{W_1, W_2, ..., W_m\}$ using the aforementioned linguistic parsing tree. Our goal is to make the attention maps of the $N_i$ attribute adjectives for the AP $W_i$, $\{W_i^j\}_{j=1}^{N_i}$, and the part noun $W_i^{N_i+1}$, \emph{i.e.}, $\{M_i^j\}_{j=1}^{N_i+1}$ share similar spatial structures. 
We therefore regard an attention map $M_i^j$ as a multi-dimensional probability distribution and define the internal structural similarity for $W_i$ as:
\begin{equation}
    d_{IS}(V_{full}, W_i) = \sum_{(j,k)\in \binom{N_i+1}{2}} d_{JS}(M_i^j, M_i^k),
\end{equation}
where $\binom{N_i+1}{2}$ denotes the 2-combination set of the $N_i+1$ indexes, $d_{JS}$ is the Jensen-Shannon Divergence~\cite{DBLP:conf/isit/FugledeT04}, and the attention mask $M_i^j$ is obtained from the cross-attention between the text token $W_i^j$ and image $V_{full}$.
We then define the semantic-bundled loss for $\{W_i\}_{i=1}^m$ as
\begin{equation}
L_{bundle}(V_{full}, \mathbf{W}) = \sum_{i=1}^m d_{IS}(V_{full}, W_i).
\end{equation}
Similarly to \cref{eq:semantic guidance}, we again shift the latent code $\hat{z_t}$ to bundle the semantics of attribute adjectives and the part noun in the APs in the denosing stage:
\begin{equation}
\label{eq:bundle guidance}
    z_t^\prime \leftarrow \hat{z_t} -\beta\cdot\nabla_{z_t}L_{bundle}(z_t, \mathbf{W}).
\end{equation}

\begin{figure*}
  \centering
  \includegraphics[width=0.97\hsize]{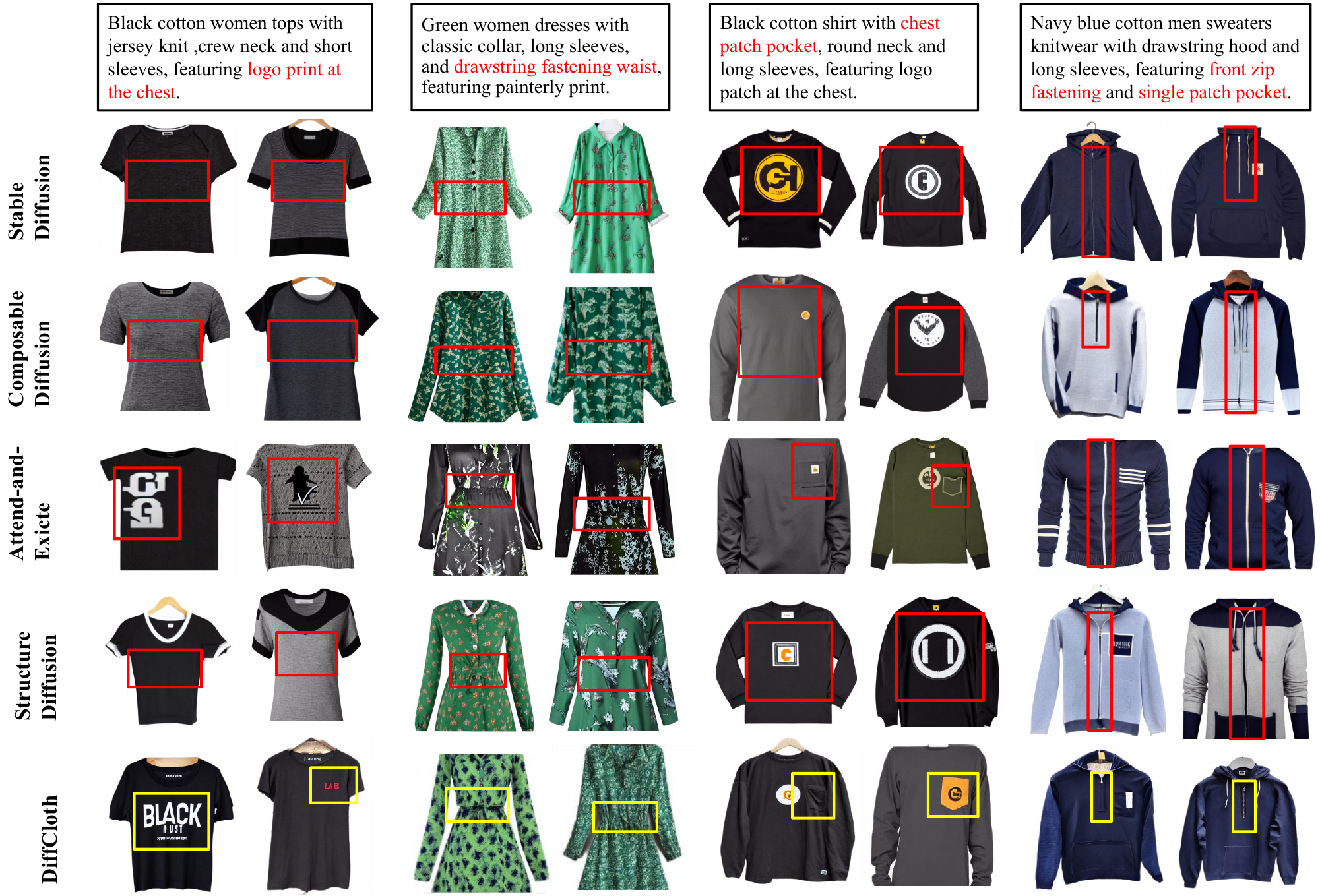}

  \caption{Results of DiffCloth on the garment synthesis task for some difficult examples that require the precise generation of fine-grained details. DiffCloth outperforms existing SOTA methods and is capable of generating semantically-correct results. The boxes are used to highlight specific areas that should contain the elements highlighted in the text.} 

\label{fig:fine}
\end{figure*}
\subsection{Region Consistency for Garment Manipulation}
\label{sec:manipulation}

DiffCloth is inspired by Prompt-to-Prompt~\cite{DBLP:journals/corr/abs-2208-01626} and allows manipulation of the generated images by simply modifying the input text prompt. Formally speaking, given an original prompt input and its $W$, we can locally manipulate an output image $I$ that is generated from $W$ by simply modifying $W$ to $W^*$ which will result in the updated image $I^*$. For example, we can change a text token $W_i^j$ to $W_i^{j,*}$ and replace its attention map $M_i^{j}$ with a new one $M_i^{j,*}$ in each diffusion step. However, we find that this simple application of Prompt-to-Prompt~\cite{DBLP:journals/corr/abs-2208-01626} degrades our bundled semantics for APs that were introduced in Sec.~\ref{sec:semantic bundle} and may lead to attribute confusion problems in the editing phase.

To preserve the bundled semantics for attribute-phrases during manipulation, as shown in Fig.~\ref{fig:edit_process}, we propose to replace the attention maps of all tokens $\{W_i^j\}_{j=1}^{N_i+1}$ in an Attribute-Phrase $W_i$ rather than solely handling the token we need to change. For example, if we want to change the attribute of the sleeves, \emph{e.g.}, ``long sleeves" $\rightarrow$ ``short sleeves", we need to inject the attention maps of both ``long" and ``sleeves". Following Prompt-to-Prompt~\cite{DBLP:journals/corr/abs-2208-01626}, we need to run the diffusion step again by merging the new attention maps $\{M_i^{j,*}\}_{j=1}^{N_i+1}$ with the fixed ones. In the $t^{th}$ denoising step, we can then use the semantic-bundled guidance in \cref{eq:bundle guidance} again to preserve the internal structural similarity for $\{M_i^{j,*}\}_{j=1}^{N_i+1}$.

Another issue with garment manipulation is how to avoid editing regions that are not relevant to the Attribute-Phrase $W_i^*$ that is being modified. To address this, we select a dynamic threshold $p$ as the first quartile of the pixel activations in the attention map $M_i^{j}$ and use it to binarize $M_i^{j}$ to a mask $B_i^j$ by thresholding. In this way, we obtain binarized masks $\{B_i^j\}_{j=1}^{N_i+1}$ and $\{B_i^{j,*}\}_{j=1}^{N_i+1}$ according to $W_i$ and $W_i^*$, respectively. The irrelevant region is then indicated by the blended mask $B_i$:
\begin{equation}
    B_i = (\bigoplus_{j=1}^{N_i+1} B_i^j) \bigoplus (\bigoplus_{j=1}^{N_i+1} B_i^{j,*}), 
\end{equation}
where $\bigoplus$ denotes boolean summation.
Similarly, when modifying multiple APs, \emph{e.g.}, $\{W_i\}_{i\in \Gamma}$, we can compute a global mask $B$ across $\{W_i\}_{i\in \Gamma}$ as $B = \bigoplus_{i\in \Gamma} B_i$,
where $\Gamma$ denotes the indexes of the APs that are being manipulated.

The region consistency is encouraged in each denoising step by blending the two latent representations $z_t$ and $z_t^*$ using $B$:
\begin{equation}
    z_{t-1}^* \leftarrow \text{Denoise}(B\cdot (z_t - z_t^*) +  z_t^*)
\end{equation}
where $\text{Denoise}(\cdot)$ denotes a DiffCloth denoising process.

\section{Experiments}
\label{sec:EXP}

\noindent \textbf{Datasets:}

Experiments are conducted on the CM-Fashion dataset~\cite{DBLP:conf/mm/ZhangSKXJHPL22}, which consists of garment images and their mask at resolution 512×512. This high-resolution fashion dataset contains 509,482 image-text pairs from various garment categories and is split into 409,482/100,000 training/testing pairs. 
In addition, we used 100,000 image-mask pairs from the training set to train the segmentor for segmenting noisy garment images into parts. 
\begin{figure}[tbp]
\centering 
\includegraphics[width=0.9\linewidth]{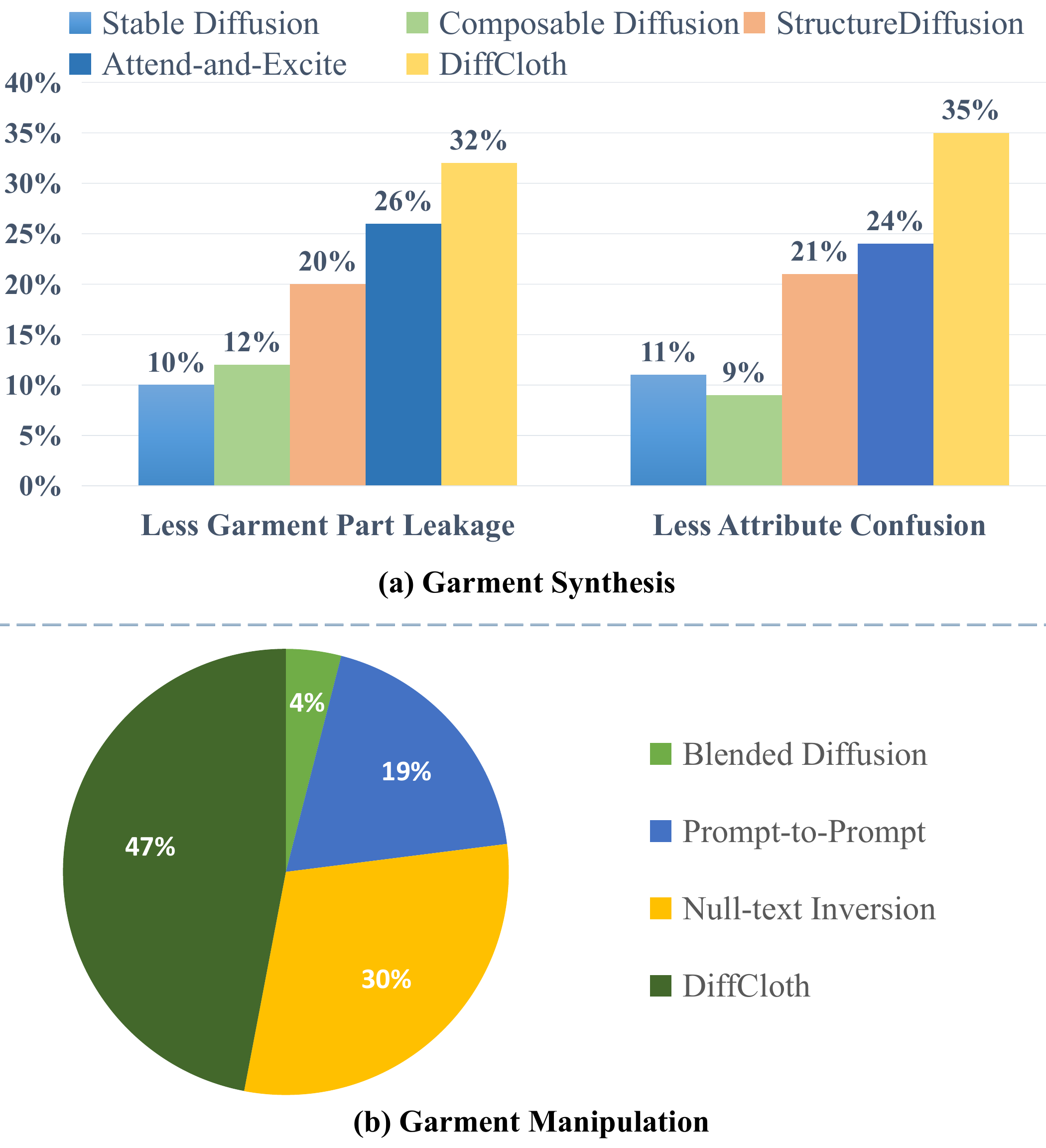} 

\caption{Human evaluation results for the garment synthesis and garment manipulation tasks.}
\label{fig:HE}

\end{figure} 
\noindent \textbf{Implementation Details:} The implementation closely follows Stable Diffusion~\cite{DBLP:conf/cvpr/RombachBLEO22}. However, we finetuned the model on the CM-Fashion dataset as the pre-trained Stable Diffusion did not produce garments on a homogeneous white background. Our models are trained on 8 Tesla V100 GPUs with a batch size of 32. During the generator training phase, the model is trained for 80 epochs with learning rate 1e-6.

Our segmentor is Pointrend~\cite{DBLP:conf/cvpr/KirillovWHG20}, which was trained using input images with added noise. The model was trained for 150 epochs with a learning rate of 4e-5. Further details are provided in the supplementary material.

\begin{table}[t]
\small
\centering
\begin{tabular}{c c c c c c}
  \toprule
  \multicolumn{2}{c}{Method}                              
  & & FID $\downarrow$ & IS $\uparrow$ & CLIPScore $\uparrow$ \\
  \cmidrule{1-2} \cmidrule{4-6} 
    \multicolumn{2}{c}{TediGAN~\cite{WeihaoXia2021TediGANTD} } 
  & & 27.37 & 18.46 & 0.5587 \\
    \multicolumn{2}{c}{Cogview~\cite{MingDing2021CogViewMT} } 
  & & 12.198 & 23.99 & 0.6572 \\
    \multicolumn{2}{c}{VQGAN~\cite{KatherineCrowson2022VQGANCLIPOD} } 
  & & 13.249 & 20.33 & 0.6423 \\
    \multicolumn{2}{c}{ARMANI~\cite{DBLP:conf/mm/ZhangSKXJHPL22} } 
  & & 12.336 & 24.32 & 0.6988 \\
  \multicolumn{2}{c}{Stable Diffusion ~\cite{DBLP:conf/cvpr/RombachBLEO22}} 
  & & 9.475 & 24.59 & 0.8169 \\
  \multicolumn{2}{c}{Composable Diffusion~\cite{DBLP:conf/eccv/LiuLDTT22}} 
  & & 9.499 & 25.91 & 0.8306 \\
  \multicolumn{2}{c}{StructureDiffusion~\cite{DBLP:journals/corr/abs-2212-05032}} 
  & & 9.238 & 25.36 & 0.8459 \\
  \multicolumn{2}{c}{Attend-and-Excite~\cite{DBLP:journals/corr/abs-2301-13826}} 
  & & 9.351 & 26.87 & 0.8241 \\
  \cmidrule{1-2} \cmidrule{4-6}
  \multicolumn{2}{c}{DiffCloth(Ours)} 
  & & \textbf{9.201} & \textbf{26.95} & \textbf{0.8974} \\  
  \bottomrule

\end{tabular}

\caption{
Comparison of DiffCloth to prior state-of-the-art approaches on the CM-Fashion dataset.}
\label{table:1}
\vspace{-5mm}
\end{table}

\noindent\textbf{Baselines and Evaluation Metrics.}
For the generation step, we compare DiffCloth to the state-of-the-art methods TediGAN~\cite{WeihaoXia2021TediGANTD}, Cogview~\cite{MingDing2021CogViewMT}, VQGAN~\cite{KatherineCrowson2022VQGANCLIPOD}, ARMANI~\cite{DBLP:conf/mm/ZhangSKXJHPL22}, Stable Diffusion~\cite{DBLP:conf/cvpr/RombachBLEO22}, Composable Diffusion~\cite{DBLP:conf/eccv/LiuLDTT22}, StructureDiffusion~\cite{DBLP:journals/corr/abs-2212-05032} and Attend-and-Excite~\cite{DBLP:journals/corr/abs-2301-13826}. To ensure fair comparisons, all models use our generator that has been trained on the CM-fashion dataset and we use the official inference code provided by the authors. For the manipulation step, we leverage Blended Diffusion~\cite{DBLP:conf/cvpr/AvrahamiLF22}, Prompt-to-Prompt~\cite{DBLP:journals/corr/abs-2208-01626}, and Null-text Inversion~\cite{RonMokady2022NulltextIF} as our primary points of comparison, as this allows us to use the same diffusion model as for DiffCloth.\footnote{Note, as Blended Diffusion~\cite{DBLP:conf/cvpr/AvrahamiLF22} is not a mask-free approach, we provide it with a manually drawn mask that reflects the text description.}
We employ three widely used metrics, namely the Fr$\mathbf{\acute{e}}$chet Inception Distance (FID)~\cite{MartinHeusel2017GANsTB}, the Inception Score (IS)~\cite{ShaneBarratt2018ANO} and the CLIPScore~\cite{JackHessel2021CLIPScoreAR} to evaluate the quality of the generation results. Furthermore, we conduct an Human Evaluation to evaluate different methods according to the text-image similarity of their results as well as their overall generation and manipulation quality.
More specifically, for the garment synthesis task, we requested that participants assess the generated images based on two criteria: the extent of garment part leakage and the amount of attribute confusion. For the garment manipulation task, we instructed them to evaluate the performance based on whether a model preserves the consistency of the content in regions that are not relevant to the manipulation.

\subsection{Comparison With State-Of-The-Art Methods}
\begin{figure}[tbp]
\centering 
\includegraphics[width=\linewidth]{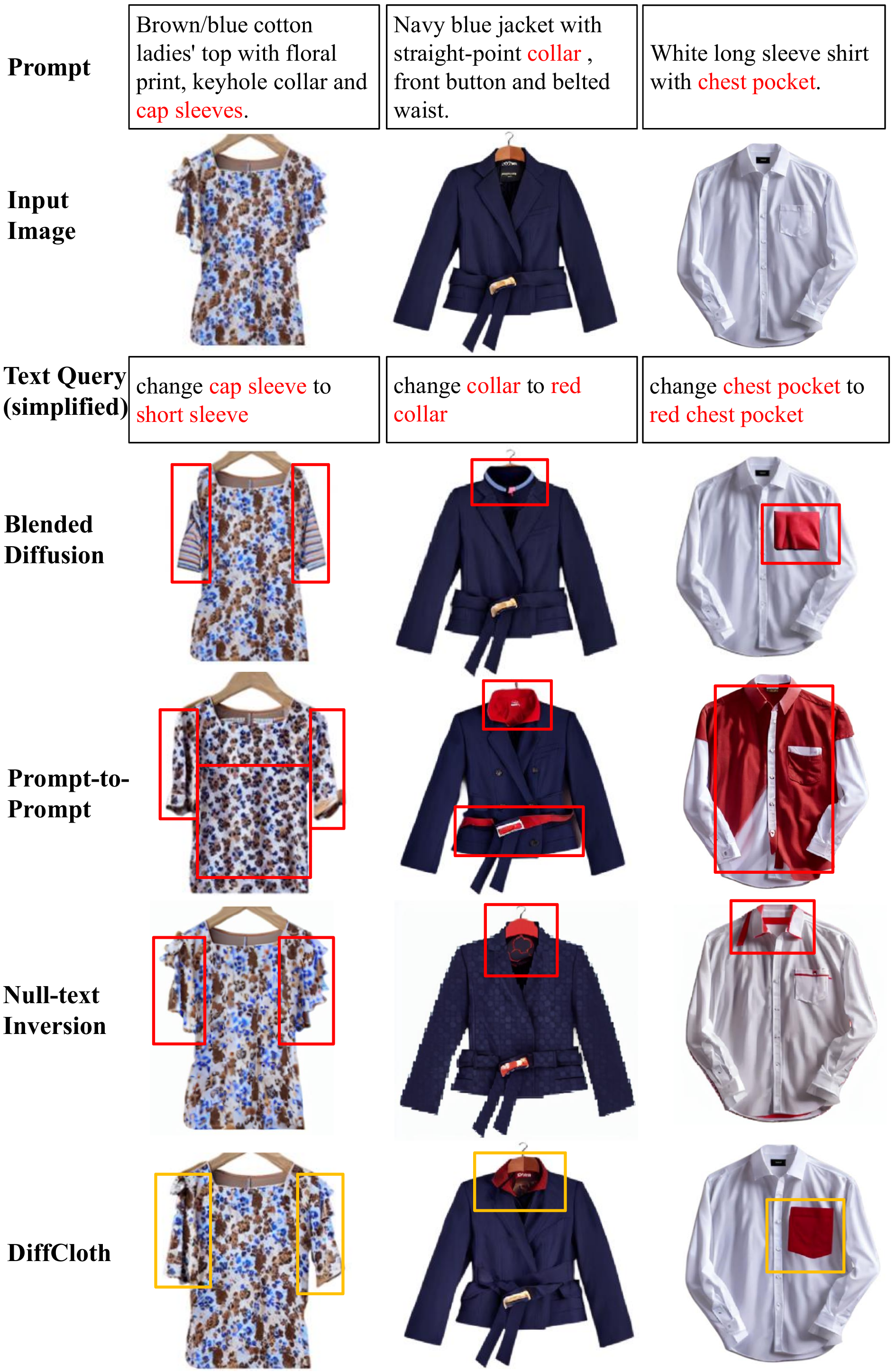} 

\caption{Results of DiffCloth for garment manipulation. The boxes are used to highlight specific areas that should contain the elements highlighted in the text.}
\label{fig:edit}
\end{figure}

\noindent\textbf{Qualitative Result}
We provide a qualitative comparison of DiffCloth's garment generation ability compared to state-of-the-art approaches~\cite{KatherineCrowson2022VQGANCLIPOD,MingDing2021CogViewMT,DBLP:journals/corr/abs-2204-06125,DBLP:conf/cvpr/RombachBLEO22}. DiffCloth is able to synthesize realistic fashion images that comply with the textual description, while prior approaches generate garment images that match the overall content of the textual description, but tends to neglect fine-grained information in the input text (red box in Fig.~{\ref{fig:fine}).
In contrast, DiffCloth is capable of generating semantically bound parts by utilizing our proposed semantic-bundled cross-attention module. Specifically, words located within an AP generate separate attributes, which enhance DiffCloth's ability to generate semantically coherent images.

For the garment manipulation, the results in Fig.~\ref{fig:edit} demonstrate the superiority of our proposed approach. We can locally manipulate an image and maintain the consistency of the content of the manipulation-irrelevant regions by using our region consistency strategy.

\textbf{Quantitative Result}
We apply FID~\cite{MartinHeusel2017GANsTB} and IS~\cite{ShaneBarratt2018ANO} to measure the quality of the synthesized images. Further, we use the CLIPScore~\cite{JackHessel2021CLIPScoreAR} to measure the relevance of the text to a given image. A higher CLIPScore indicates that the text is more relevant to the image. As reported in Tab.~\ref{table:1}, our proposed DiffCloth outperforms the baselines Stable Diffusion~\cite{DBLP:conf/cvpr/RombachBLEO22}, Composable diffusion~\cite{DBLP:conf/eccv/LiuLDTT22}, StructureDiffusion~\cite{DBLP:journals/corr/abs-2212-05032} and Attend-and-Excite~\cite{DBLP:journals/corr/abs-2301-13826} in all cases by a large margin, obtaining the lowest FID score and the highest IS and CLIPScore for the garment synthesis.
In addition, we designed two human evaluation studies to quantitatively compare the generation and manipulation results with the baselines. For generation, we ask participants to select the generated results that exhibit minimal attribute confusion and Garment Part Leakage. For the manipulation task, we evaluate the effectiveness of the method by asking participants to select the results that best preserves the area that is irrelevant to the text modification. Aggregating the scores per model in Fig.~\ref{fig:HE}, we observe that DiffCloth's results are preferred for both the garment synthesis or manipulation tasks. Furthermore, it is also noticeable that the human-based evaluation indicates a larger difference among the models compared to the machine evaluation.

\subsection{Ablation study}

\begin{table}[t]
\small
\centering
\begin{tabular}{l c c c c c c c c }
  \toprule
  \multicolumn{2}{c}{Method}                              
  & & L1 & L2 & & FID$\downarrow$  & IS $\uparrow$ & CLIPScore $\uparrow$ \\
  \cmidrule{1-2} \cmidrule{4-5} \cmidrule{7-9} 
  \multicolumn{2}{c}{DiffCloth$\dagger$} 
  & & \xmark & \xmark & & 9.475 & 24.59 & 0.8169\\
  
  \multicolumn{2}{c}{DiffCloth$\star$} 
  & & \cmark & \xmark & & 9.381 & 25.45 & 0.8821\\
  
  \multicolumn{2}{c}{DiffCloth$\ast$} 
  & & \xmark & \cmark & & 9.221 & 26.69 & 0.8423 \\  
  \cmidrule{1-2} \cmidrule{4-5} \cmidrule{7-9} 
  \multicolumn{2}{c}{DiffCloth} 
  & & \cmark & \cmark  & & \textbf{9.201} & \textbf{26.95} & \textbf{0.8974} \\ 
  \bottomrule
\end{tabular}
\caption{Quantitative results of our ablation studies. L1 and L2 denote the structural semantic consensus guidance and the semantic-bundled cross-attention, respectively.}

\label{table:2}
\end{table}
In the garment synthesis task, to validate the effectiveness of the structural semantic consensus guidance and the semantic-bundled cross-attention, we design three variants of our proposed method and evaluate the performance of the different variants according to their metric scores. We denote Stable Diffusion~\cite{DBLP:conf/cvpr/RombachBLEO22} as DiffCloth$\dagger$, DiffCloth without structural semantic consensus guidance as DiffCloth$\star$, and denote DiffCloth without semantic-bundled cross-attention as DiffCloth$\ast$.
For the garment manipulation task, we consider DiffCloth without region consistency as our ablated model and denote it as DiffCloth$\pounds$.

\begin{figure}[tbp]
\centering 
\includegraphics[width=\linewidth]{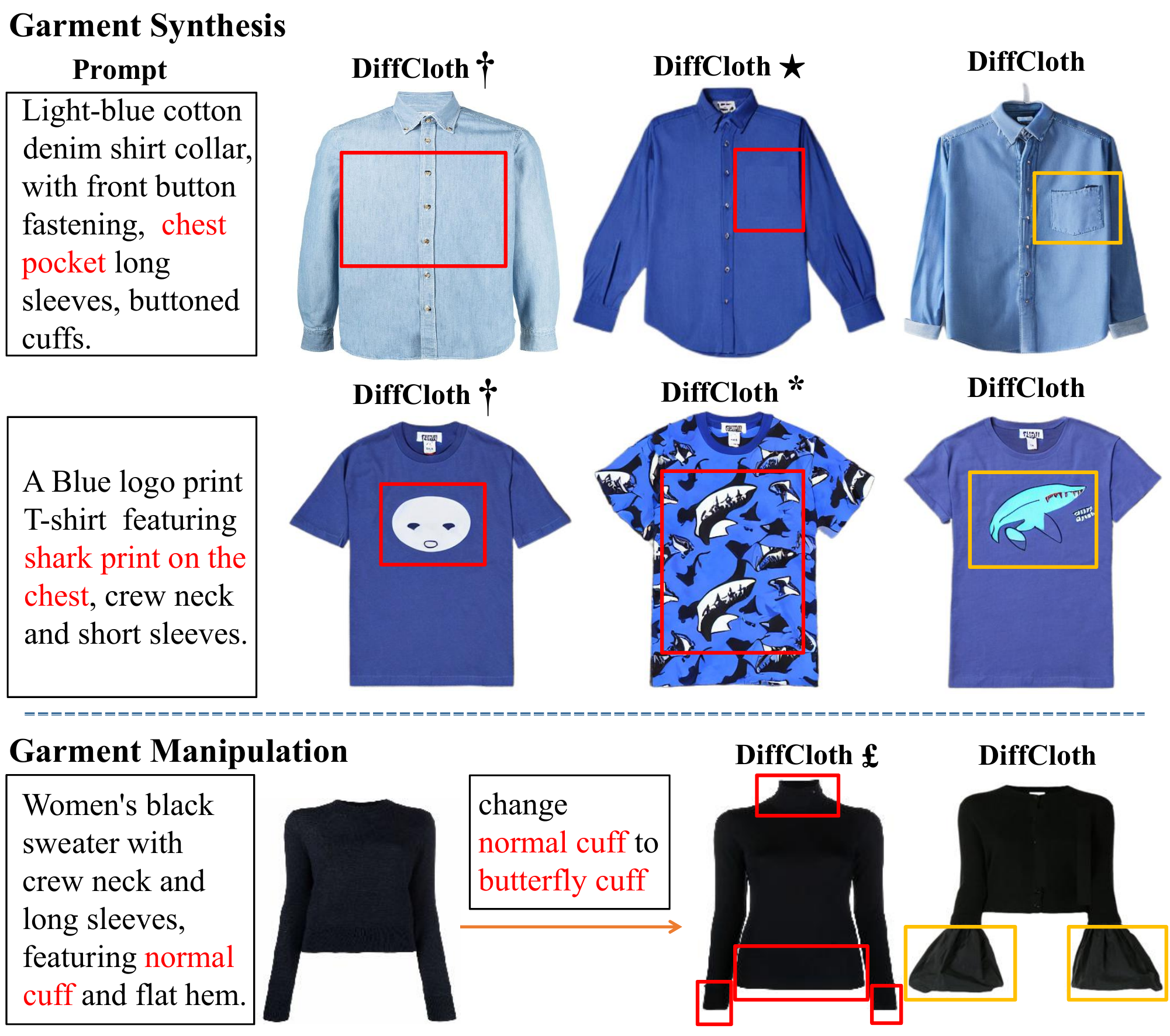}

\caption{Qualitative results of our ablation studies for garment synthesis (top) and manipulation (bottom).}

\label{fig:ablation}
\end{figure}

As reported in Tab.~\ref{table:2}, incorporating either the structural semantic consensus guidance or the semantic-bundled cross-attention (or both) leads to significant improvements in FID, IS and CLIPScore. These results indicate that our proposed mechanisms can produce more realistic and semantically accurate results. Additionally, as illustrated in Fig.~\ref{fig:ablation}, the incorporation of structural semantic consensus guidance (as DiffCloth$\star$) leads to the generation of more accurate parts, whereas the exclusion of the semantic-bundled cross-attention increases attribute confusion. Finally, removing the region consistency strategy in garment manipulation causes the model to affect parts that should not be modified, as demonstrated in Fig.~\ref{fig:ablation}.

\section{Conlusion}
In this work, we propose DiffCloth, a diffusion-based pipeline for garment synthesis and manipulation, which aligns the structural cross-modal semantics between input prompts and garment images to address the problem of garment part leakage and attribute confusion. Moreover, DiffCloth provides a convenient way to manipulate its generated garments by replacing the Attribute-Phrase in the text prompt, while ensuring that the content in regions unrelated to the modification is preserved using a consistency loss. Experiments on the CM-Fashion demonstrate DiffCloth's superior effectiveness compared to existing methods.

\noindent\textbf{Limitation and future work:} A limitation of our approach is the sensitivity to noisy text, which may make accurate correspondance matching more challenging. To address this limitation, we aim to explore how the text information can be leveraged to further strengthen the model's robustness. 

\section{Acknowledgement}

\looseness=-1
This work was supported in part by National Key R$\&$D Program of China under Grant No.2020AAA0109700,
Guangdong Outstanding Youth Fund(Grant No.2021B1515020061),
Shenzhen Science and Technology Program(Grant No.RCYX20200714114642083), Shenzhen Fundamental Research Program(Grant No.JCYJ20190807154211365),
Nansha Key RD Program under Grant No.2022ZD014 and Sun Yat-sen University under Grant No.22lgqb38 and 76160-12220011. We thank MindSpore for the partial support of this work, which is a new deep learning computing framwork.\footnote{https://www.mindspore.cn}.

{\small
\bibliographystyle{ieee_fullname}
\bibliography{egbib}
}

\end{document}